\documentclass[conference]{IEEEtran}
\IEEEoverridecommandlockouts

\usepackage{cite}
\usepackage{amsmath,amssymb,amsfonts}
\usepackage{algorithmic}
\usepackage{graphicx}
\usepackage{textcomp}
\usepackage{xcolor}
\usepackage{multirow}
\usepackage{booktabs}
\usepackage{url}
\usepackage{float}
\usepackage{subcaption}
\def\BibTeX{{\rm B\kern-.05em{\sc i\kern-.025em b}\kern-.08em
    T\kern-.1667em\lower.7ex\hbox{E}\kern-.125emX}}
\begin{document}

\title{CASPER: A Large Scale Spontaneous Speech Dataset\\
}

\author{
\IEEEauthorblockN{Cihan Xiao}
\IEEEauthorblockA{
Johns Hopkins University\\
Baltimore, USA\\
cxiao7@jhu.edu}
\and
\IEEEauthorblockN{Ruixing Liang}
\IEEEauthorblockA{
Johns Hopkins University\\
Baltimore, USA\\
rliang7@jhu.edu}
\and
\IEEEauthorblockN{Xiangyu Zhang}
\IEEEauthorblockA{Johns Hopkins University\\
Baltimore, USA\\
xiangyu.zhang2@unsw.edu.au}
\and
\IEEEauthorblockN{Mehmet Emre Tiryaki}
\IEEEauthorblockA{
Johns Hopkins University\\
Baltimore, USA\\
mtiryak1@jhu.edu}
\and
\IEEEauthorblockN{Veronica Bae}
\IEEEauthorblockA{
Johns Hopkins University\\
Baltimore, USA\\
vbae1@jhu.edu}
\and
\IEEEauthorblockN{Lavanya Shankar}
\IEEEauthorblockA{
Johns Hopkins University\\
Baltimore, USA\\
ls1@jhu.edu}
\and
\IEEEauthorblockN{Rong Yang}
\IEEEauthorblockA{
Johns Hopkins University\\
Baltimore, USA\\
ryang42@jhu.edu}
\and
\IEEEauthorblockN{Ethan Poon}
\IEEEauthorblockA{Edison Academy Magnet School\\
USA\\
ewpoon007@gmail.com}
\and
\IEEEauthorblockN{Emmanuel Dupoux}
\IEEEauthorblockA{Meta\\
USA\\
dpx@meta.com}
\and
\IEEEauthorblockN{Sanjeev Khudanpur}
\IEEEauthorblockA{
Johns Hopkins University\\
Baltimore, USA\\
khudanpur@jhu.edu}
\and
\IEEEauthorblockN{Leibny Paola Garcia Perera}
\IEEEauthorblockA{
Johns Hopkins University\\
Baltimore, USA\\
lgarci27@jhu.edu}
}



\maketitle

\begin{abstract}
The success of large language models has driven interest in developing similar speech processing capabilities. However, a key challenge is the scarcity of high-quality spontaneous speech data, as most existing datasets contain scripted dialogues. To address this, we present a novel pipeline for eliciting and recording natural dialogues and release our dataset with 100+ hours of spontaneous speech. Our approach fosters fluid, natural conversations while encouraging a diverse range of topics and interactive exchanges. Unlike traditional methods, it facilitates genuine interactions, providing a reproducible framework for future data collection. This paper introduces our dataset and methodology, laying the groundwork for addressing the shortage of spontaneous speech data. We plan to expand this dataset in future stages, offering a growing resource for the research community.

\end{abstract}


\section{Introduction}
Recent advances in large language models have demonstrated unprecedented capabilities in natural language understanding and generation~\cite{touvron2023llama,dubey2024llama}. These models have achieved remarkable performance across various tasks, from text completion to complex reasoning, fundamentally transforming how we approach natural language processing challenges~\cite{singh2025prompted,chen2023soulchat,zhang2024llms}. The success of these models lies not only in their architectural innovations but also in their ability to learn from vast amounts of diverse, high-quality text data. This success has naturally led researchers to explore similar architectures for speech processing, aiming to develop models that can understand and generate human speech with comparable sophistication~\cite{chu2023qwen,du2024cosyvoice,du2024cosyvoice2}. The potential applications of such speech language models are extensive, ranging from more natural human-computer interaction to advanced speech translation and synthesis systems.

However, the development of powerful speech language models faces a fundamental challenge that distinguishes it from traditional text-based approaches: the limited availability of high-quality spontaneous speech data. While text corpora can be readily collected from various online sources, gathering natural conversational speech presents unique difficulties. Existing speech datasets predominantly consist of scripted dialogues or read speech, which often fail to capture the nuanced characteristics of natural conversation, including hesitations, overlaps, and authentic turn-taking behaviors~\cite{panayotov2015librispeech,chen2021gigaspeech}. This limitation has become a critical bottleneck in advancing speech processing technologies, particularly in applications requiring understanding and generation of natural conversational speech.

To address this crucial gap, we introduce a comprehensive speech dataset named \textbf{CA}sual \textbf{S}peech in \textbf{P}eer \textbf{E}ngagement \textbf{R}ecordings, or \textbf{CASPER}, with 200 hours of recorded English conversations, of which 158 hours are actual speech. We collected a total of 200 hours of speech; however, this release contains the first 102 hours to support early research and development.\footnote{ \url{https://huggingface.co/datasets/CASPER-SSSD/CASPER/} } The remaining data will be released in conjunction with a future challenge to encourage research involvement and participation.
This subset of the dataset is partitioned into train (80\%) and test (20\%) splits. To ensure the integrity of the evaluation, we release speaker metadata (e.g., accent, age) and labels for both splits. This is an initial effort to expand the dataset to 1,000 hours of natural, spontaneous dialogue, enriched with metadata such as speaker demographics and recording devices. 

Central to this research is the investigation of interactions in typical telephone conversations. Our dataset captures the spontaneity, turn-taking dynamics, and informal speech patterns essential for improving downstream tasks such as Automatic Speech Recognition (ASR), speaker recognition, diarization, and topic detection performance in real-world conversational settings. Our data collection platform specifically targets \emph{pairs of individuals who already know each other}, fostering natural, unforced conversations that can capture the properties of real-world conversations.

The pipeline's device-agnostic architecture allows for recording across various hardware setups, from professional audio equipment to common mobile devices, making it highly accessible for future data collection efforts. In this paper, we present our initial release, which comprises a subset of 102 hours from a total of 200 hours of spontaneous speech, along with partial transcriptions. Unlike existing datasets such as Switchboard~\cite{godfrey1992switchboard}, which primarily features shorter conversations and limited recording conditions, or GigaSpeech~\cite{chen2021gigaspeech}, which relies on existing media sources, our approach actively facilitates extended, natural dialogues that better capture the full complexity of human conversation. The longer conversation duration enables the development of more complex topics and natural interaction patterns, while our multi-device support ensures consistent quality across different recording environments. By releasing both the initial dataset and our flexible collection framework, we aim to establish a foundation for addressing the scarcity of spontaneous speech data, enabling researchers to study various aspects of natural conversation while contributing to this growing resource.

\section{Related Work}
\subsection{Speech Large Language Model (Speech LLM)}
Recent years have witnessed significant progress in developing large language models for speech processing tasks. Building upon the success of text-based LLMs, researchers have explored various approaches to extend language model capabilities to speech understanding and generation. Initial efforts focused on combining existing LLMs with speech encoders~\cite{chu2023qwen}, while more recent work has attempted to train speech-to-speech language models~\cite{zhang2023speechgpt,nguyen2023generative,du2024cosyvoice,borsos2023audiolm,lakhotia2021generative}. Notably, models like CosyVoice~\cite{du2024cosyvoice} and CosyVoice2~\cite{du2024cosyvoice2} have demonstrated the potential of unified architectures for multiple speech tasks. Moshi~\cite{defossez2024moshi} further advanced this direction by introducing a large-scale speech foundation model capable of both understanding and generating speech content. These models, however, primarily rely on carefully curated speech datasets or modified text-based pre-training approaches, highlighting the need for large-scale, high-quality spontaneous speech data for training more capable speech language models and evaluating the current model with more real-life challenges.

\subsection{Existing Conversational Speech Corpora}

Several datasets have been created to capture conversational speech, each with its own focus and limitations. The Switchboard Telephone Speech Corpus~\cite{godfrey1992switchboard}, collected in the early 1990s, contains 260 hours of telephone conversations from 543 speakers across the United States. While groundbreaking for its time, its conversations were relatively short and limited by telephone audio quality.
Similar to Switchboard, the CALLHOME American English Speech corpus~\cite{LDC97T14} also utilized telephone conversations but faced similar limitations in conversation length and audio fidelity.

The Santa Barbara Corpus of Spoken American English (SBCSAE)~\cite{du2000santa} took a different approach, focusing on capturing naturally occurring interactions across diverse settings. With approximately 249,000 words of transcribed speech, it includes recordings from various scenarios such as card games, food preparation, classroom lectures, and town hall meetings. While this diversity is valuable, the uncontrolled recording environments often result in inconsistent audio quality. The HCRC Map Task Corpus~\cite{anderson1991hcrc} approached conversational speech collection through a structured task-based framework, recording 128 unscripted dialogues where participants collaborated on map-related tasks. However, this task-oriented approach may not fully reflect natural conversation patterns.

More recently, Meta's Casual Conversation Dataset~\cite{porgali2023casual} represents a modern approach to conversational data collection, capturing both video and audio from 3,011 subjects. While this dataset primarily focuses on supporting computer vision and audio model evaluation, with annotations for demographic attributes and recording conditions, it doesn't specifically target extended natural conversations or address the needs of accent-diverse speech processing applications.

Also being part of the Scalable Spontaneous Speech Dataset (SSSD) project, \cite{sheikh24_interspeech} adopts a crowdsourced approach to spontaneous speech collection. Like our methodology, they utilize a web-based platform for data collection, ensuring accessibility and ease of use. While their pipeline relies on Amazon Mechanical Turk (MTurk) for participant recruitment, often involving strangers, whereas our approach avoids crowdsourcing marketplaces in favor of more targeted recruitment, specifically acquaintances. This allows us to foster more natural interactions, reducing the risk of rushed or disengaged contributions and resulting in more meaningful and coherent dialogues.

\section{Data Collection}
\subsection{Data Collection Procedure}
Our dataset comprises audio recordings of spontaneous conversations between pairs of participants, captured in real-world settings using their personal devices. To facilitate a structured yet natural data collection process, we developed a web-based platform where participants could initiate and record their conversations.\footnote{We open-source our code at https://anonymous.4open.science/r/sssd-D5CA/README.md.}

\begin{figure}[htbp]
  \centering
    \includegraphics[width=0.5\textwidth, clip]{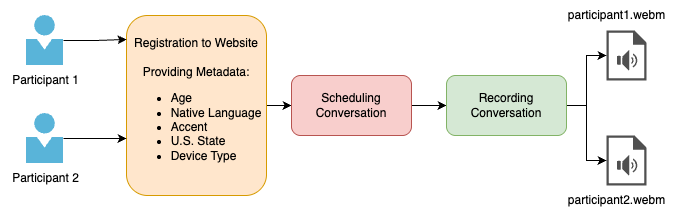}
  \caption{Participant recording pipeline.}

  \label{fig:flowchart}
\end{figure}

Participants were recruited through emails and flyers, after which they registered on the platform, verified their email, and provided metadata, including age, native language, accent, and device type. Prior to participation, they reviewed and signed an informed consent form. A visualization of the pipeline is shown in Fig. \ref{fig:flowchart}. Pairs of peer participants already familiar with each other were targeted in the recruitment process to facilitate natural conversations. To prioritize the study of natural, spontaneous conversations, our dataset places secondary emphasis on diversity in gender, language, and other demographic factors.

\begin{figure}[htbp]
  \centering
    \includegraphics[width=0.35\textwidth, clip]{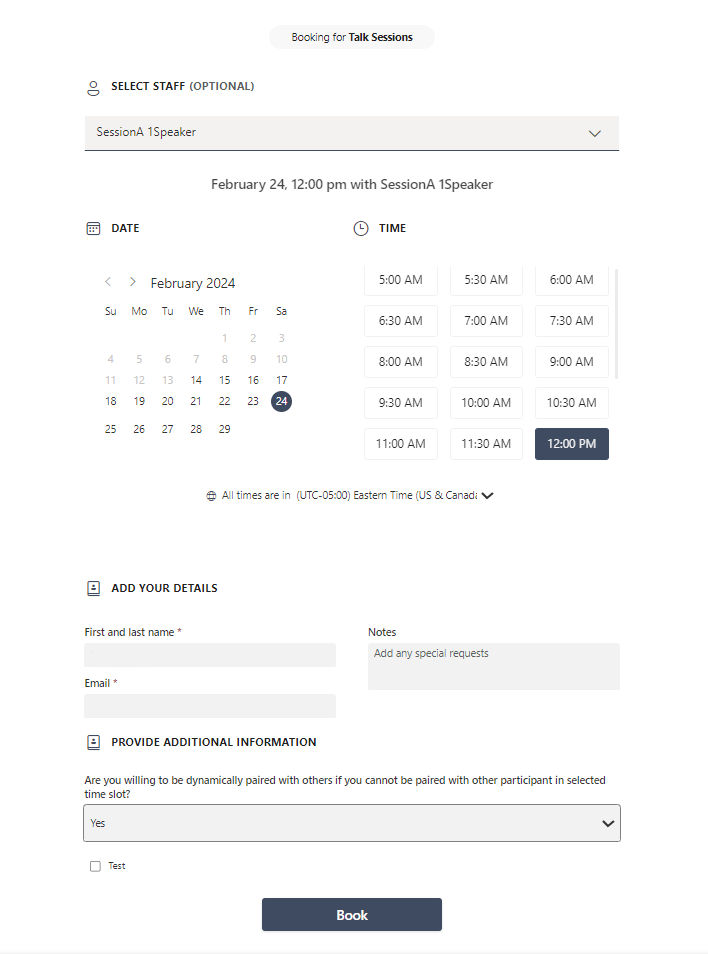}
  \caption{Recording Scheduling Tool.}

  \label{fig:scheduler}
\end{figure}

Each participant was able to schedule recordings per their own timeline using the developed scheduling tool seen in Fig. \ref{fig:scheduler}. 

\begin{figure}[htbp]
  \centering
    \includegraphics[width=0.5\textwidth, clip]{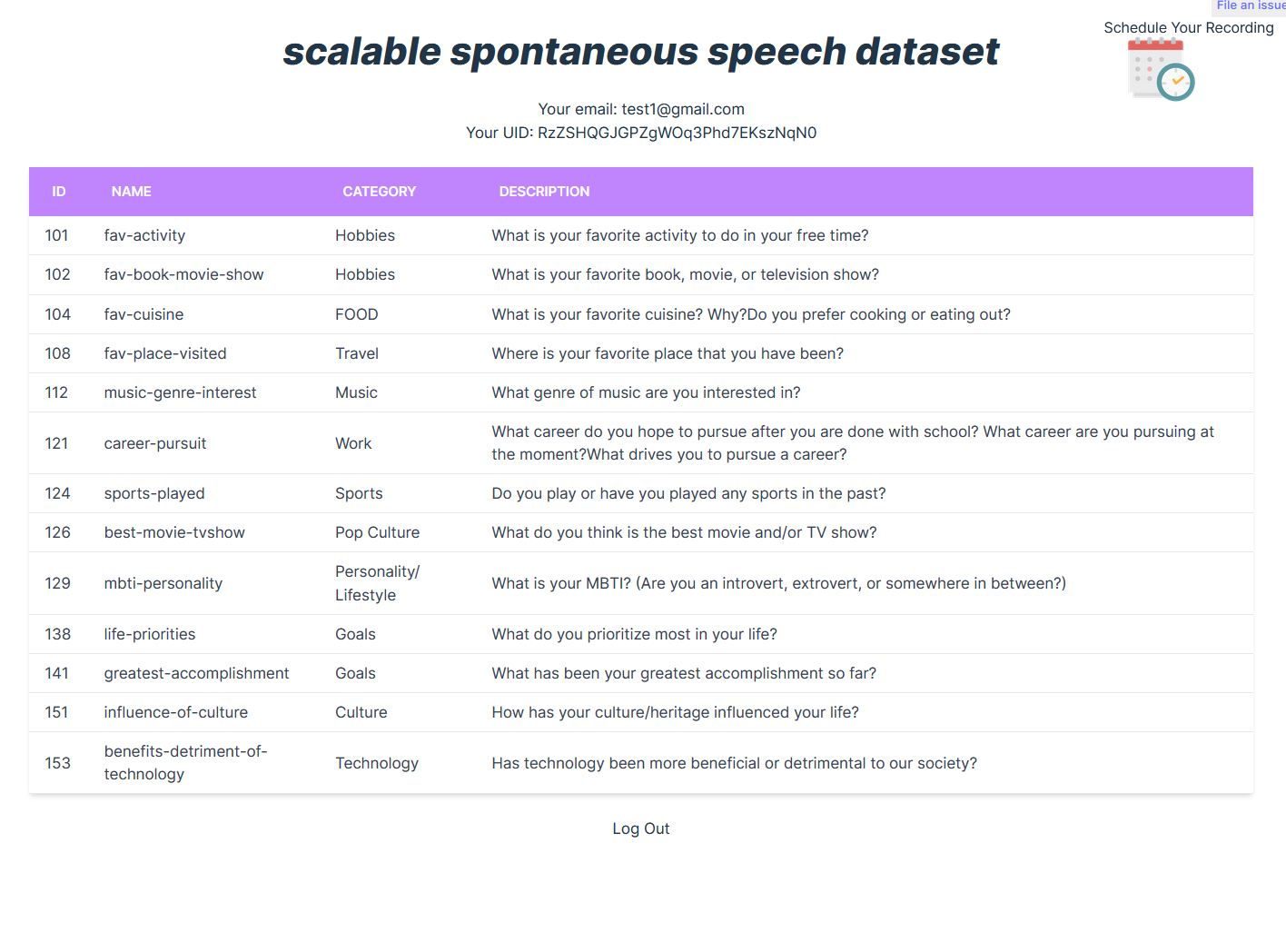}
  \caption{Topic Selection Screen.}

  \label{fig:topics}
\end{figure}

To encourage natural dialogue, participants selected conversation rooms labeled with broad topic descriptors (e.g., Favorite Activities) seen in Fig. \ref{fig:topics}. 

\begin{figure}[htbp]
  \centering
  \includegraphics[width=0.5\textwidth, clip]{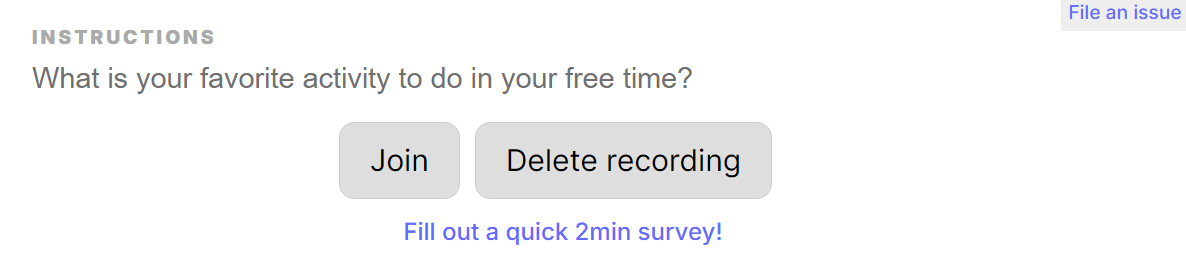}
  \caption{Recording Screen.}
  \label{fig:recording}

  \vspace{1em} 

  \includegraphics[width=0.25\textwidth, clip]{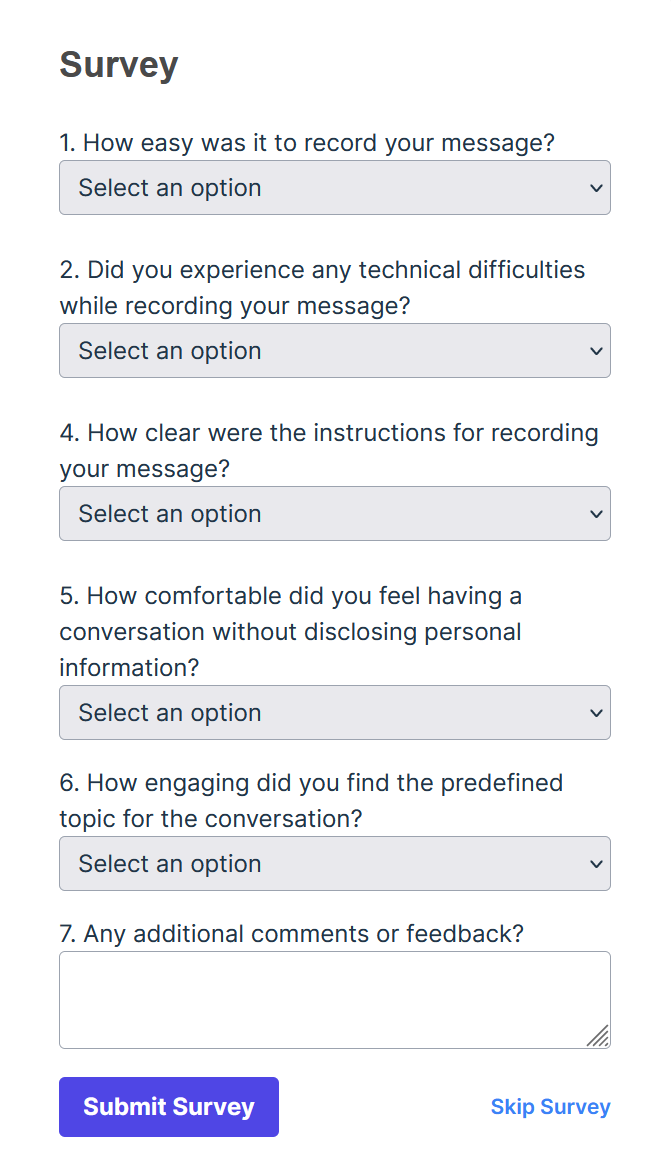}
  \caption{Post-Recording Survey.}
  \label{fig:survey}
\end{figure}

These prompts served as conversation starters rather than strict guidelines, allowing discussions to evolve freely. Each participant’s audio was recorded in a separate channel for enhanced signal quality and post-processing flexibility. Upon completion, participants provided feedback through a survey assessing the ease and quality of the recording process. The recording screen from the participants' end and the post-recording survey can be seen in Fig. \ref{fig:recording} and Fig. \ref{fig:survey} respectively. 

This methodology ensures a scalable, reproducible framework for collecting high-quality spontaneous speech, with rich metadata that enhances the dataset’s utility for research in speech processing.

\subsection{Data Collection Format}
The conversations were recorded using the backend infrastructure of \textit{Daily.co}\footnote{\url{https://www.daily.co/}}. Each participant’s audio, recorded in a separate channel, enables downstream processing tasks such as automatic speech recognition per-speaker, speaker diarization and analysis of overlapping speech. This per-channel recording approach enhances the dataset’s utility for speech processing research by providing cleaner, source-separated signals rather than a single mixed audio stream.
The recorded audio files were initially stored in an \textit{Amazon AWS S3 bucket} in \textit{WebM} format with a sample rate of 48 kHz. To ensure long-term accessibility and compatibility with standard speech processing tools, the files were subsequently transferred to an internal computing cluster and converted to \textit{WAV} format at the same sample rate, with no loss in fidelity.

User metadata was exported from the \textit{Firebase}\footnote{\url{https://firebase.google.com/}} Realtime Database in \textit{JSON} format. In order to protect sensitive information, personally identifiable data (e.g., names and geographic details) were stored separately, indexed by a unique User ID, and restricted to administrative access only. 

\subsection{Data Storage and Post-Processing}
To facilitate indexing and statistical analysis, the \textit{WAV} recordings and their corresponding \textit{JSON} metadata were integrated into a \textit{SQL} database. Next, voice activity detection (VAD) using Silero-v4 \cite{Silero} and ASR using Whisper-large-v3 \cite{radford2022robustspeechrecognitionlargescale} were performed to estimate the ratio of speech to nonspeech for each recording, revealing that speech comprised about 78\% of the audio. The dataset was then refined using Label Studio\footnote{\url{https://labelstud.io/}}—an open-source annotation platform—for human annotation, enabling fine-grained labeling of high-quality transcripts. 

\subsection{De-Identification}

To protect the privacy of the participants the audio files, as well as the metadata, were de-identified. To achieve this, each recording was transcribed along with word-level timestamps using WhisperX \cite{bain2022whisperx}. After this, each transcription was searched to find any occurrence of the participant provided metadata. The identified metadata information, such as names and surnames, were then replaced with silences in the recordings based on their word-level timestamps.

\section{Data Overview}
\subsection{Metadata Analysis}
The dataset comprises a total of 208 participants, each contributing spontaneous speech recordings along with associated metadata, as shown in Table~\ref{tab:metadata}. The inclusion of demographic and technical details enhances the dataset’s utility, providing researchers with the necessary context to analyze speaker characteristics, recording conditions, and linguistic diversity. Below, we discuss the distribution of key metadata attributes in detail.

\subsubsection{Speaker Demographics}
The participants exhibit a diverse range of linguistic backgrounds and regional accents. The majority (67.79\%) reported speaking US English, reflecting the dataset’s primary demographic. However, a significant proportion of non-native and regionally influenced English varieties are also present, including Chinese Mandarin-influenced English (4.81\%), UK English (5.29\%), and Indian English (2.88\%). Additionally, 14.42\% of participants did not specify an accent, indicating either an omission or variability in self-identification. The participants’ accent and native language are based on their self-identification, for example, the number of speakers with an Arabic accent may differ from the number with Arabic as their native language.

Age distribution reveals that younger speakers are overrepresented, with 57.21\% of participants in the 18-29 age range and 23.56\% in the 30-39 range. Older age groups are comparatively sparse, with only 0.48\% aged 50-59 and 60-69. This skew towards younger speakers may reflect recruitment channels and the tendency of younger individuals to engage with online data collection platforms.

In terms of gender, 48.56\% of participants identified as male, while 24.04\% identified as female. Additionally, a small fraction (1.44\%) is identified as non-binary, contributing to the dataset’s inclusivity. Although the number of male speakers was higher, the total speech duration between male and female participants remained comparable (35.45\% vs. 37.09\%), indicating balanced contributions in terms of spoken content.

\subsubsection{Linguistic Diversity}
The dataset features a majority of native English speakers (71.15\%), but it also includes a range of other native languages. Chinese (6.25\%) and Spanish (1.92\%) are among the most represented, followed by Arabic (0.96\%), Hindi (0.96\%), and a variety of other languages. The presence of non-native speakers introduces variations in pronunciation, fluency, and code-switching patterns, making the dataset particularly valuable for studies in accented speech recognition.


\begin{table}[t]
\centering
\caption{Metadata distribution of the collected dataset over the total 208 distinct speakers. Absolute counts and percentages are provided for each attribute. \textbf{NL} refers to the native language of the speaker}
\begin{tabular}{l c c c c}
\toprule
\textbf{Attribute} & \textbf{Category} & \textbf{Count} & \textbf{Duration} \\
& & (\#) / (\%) & (Sec) / (\%) \\
\midrule
\multirow{6}{*}{\textbf{Accent}} 
            & US English & 141 / 67.79\% & 402,970 / 70.66\% \\
            & UK English & 11 / 5.29\% & 22,599 / 3.96\% \\
            & CMN English & 10 / 4.81\% & 13,363 / 2.34\% \\
            & Indian English & 6 / 2.88\% & 4,020 / 0.70\% \\
            & N/A & 30 / 14.42\% & 97,636 / 17.12\% \\
            & Others & 10 / 4.81\% & 29,669 / 5.20\% \\  
        \midrule
        \multirow{6}{*}{\textbf{Age}} 
            & 18-29 & 119 / 57.21\% & 270,262 / 47.39\% \\
            & 30-39 & 49 / 23.56\% & 180,446 / 31.64\% \\
            & 40-49 & 1 / 0.48\% & 732 / 0.13\% \\
            & 50-59 & 1 / 0.48\% & 3,457 / 0.61\% \\
            & 60-69 & 1 / 0.48\% & 267 / 0.05\% \\
            & N/A & 37 / 17.79\% & 115,143 / 20.19\% \\
        \midrule
        \multirow{6}{*}{\textbf{Device}} 
            & iPhone & 92 / 44.23\% & 353,396 / 61.97\% \\
            & Android & 38 / 18.27\% & 47,744 / 8.37\% \\
            & MacBook & 15 / 7.21\% & 22,165 / 3.89\% \\
            & PC & 6 / 2.88\% & 4,869 / 0.85\% \\
            & iPad & 2 / 0.96\% & 5,420 / 0.95\% \\
            & Others / N/A & 55 / 26.44\% & 136,713 / 23.97\% \\
        \midrule
        \multirow{4}{*}{\textbf{Gender}} 
            & Male & 101 / 48.56\% & 202,171 / 35.45\% \\
            & Female & 50 / 24.04\% & 211,515 / 37.09\% \\
            & Non-binary & 3 / 1.44\% & 2,568 / 0.45\% \\
            & N/A & 54 / 25.96\% & 154,053 / 27.01\% \\
        \midrule
        \multirow{6}{*}{\textbf{NL}} 
            & English & 148 / 71.15\% & 427,058 / 74.88\% \\
            & Chinese & 13 / 6.25\% & 19,739 / 3.46\% \\
            & Spanish & 4 / 1.92\% & 9,266 / 1.62\% \\
            & Arabic & 2 / 0.96\% & 3,003 / 0.53\% \\
            & Hindi & 2 / 0.96\% & 1,225 / 0.21\% \\
            & Others / N/A & 39 / 18.75\% & 110,016 / 19.30\% \\

\bottomrule
\end{tabular}
\label{tab:metadata}
\end{table}

  

\begin{table*}[!htbp]
    \centering
    \renewcommand{\arraystretch}{1.3} 
        \caption{Example of spontaneous dialogue with disfluencies and long-context dependencies.}
    \label{tab:spontaneous_speech}
    \begin{tabular}{p{2cm} p{12cm}}  
        \toprule
        \textbf{Speaker} & \textbf{Utterance} \\  
        \midrule
        Spk1 & [um], I was kind of, [um], having, [um], network issues, so I have been trying to resolve that. \\  
        Spk2 & No, that's very okay. \\  
        \multicolumn{2}{c}{[... long conversation ...]} \\  
        Spk2 & Yeah, so, me, I'm not, I'm not someone that really travels a lot. \\  
        \multicolumn{2}{c}{[... long conversation ...]} \\  
        Spk1 & And just like you I don't, [cough], I don't really, [um], travel a lot. \\  
        \multicolumn{2}{c}{[... long conversation ...]} \\  
        Spk1 & Cause I kind of, [um], spend a little bit of my, [um], little age growing up here, so \dots \\  
        \bottomrule
    \end{tabular}
\end{table*}

\subsubsection{Recording Conditions and Device Usage}
The dataset was collected using a wide range of recording devices, reflecting real-world variability in speech capture conditions. The most common devices were iPhones (44.23\%) and Android smartphones (18.27\%), followed by MacBooks (7.21\%) and PCs (2.88\%). A smaller subset of participants recorded using iPads (0.96\%) or dedicated audio peripherals such as headsets. This diversity in hardware introduces natural variations in microphone quality, background noise levels, and acoustic artifacts, which are crucial for training robust speech recognition models.

\subsubsection{Post-recording Survey Analysis}
To evaluate the user experience, we conducted a post-recording survey with 181 participants. Among them, 143 (79.01\%) reported that recording the conversations was "very easy", indicating that our data collection pipeline was well-designed, intuitive, and minimally intrusive. Furthermore, 136 participants (75.14\%) found their conversation topics "very engaging", suggesting that our approach successfully fosters natural and meaningful discussions. These results highlight the strength of our data collection methodology, demonstrating that most participants were not only willing but also comfortable and engaged while recording.

\subsection{Data Description}
\subsubsection{Overview}
Our dataset consists of a total of 200 hours of recorded audio, with 158 hours identified as actual speech based on our VAD results. To maintain meaningful interactions, we filtered out conversations shorter than 10 seconds, focusing on those that provide substantial linguistic content. The longest conversation in our dataset lasted approximately 1.89 hours (6,819.6 seconds), while the mean and median conversation lengths are 948.2 seconds (15.8 minutes) and 913.6 seconds (15.2 minutes), respectively. The recordings capture spontaneous, conversational speech, featuring natural disfluencies, hesitations, and turn-taking patterns crucial for real-world ASR. This distribution reflects our emphasis on capturing extended, interactive dialogues, which better emulate natural telephone conversations and real-world speech dynamics. Unlike crowdsourced datasets, a significant portion of conversations occurred between acquaintances, fostering more natural interactions.
\subsubsection{Naturalness}
As shown in Table~\ref{tab:spontaneous_speech}, the transcript reflects key characteristics of spontaneous, natural conversations. Unlike scripted speech, natural dialogues often contain disfluencies such as fillers ("um"), restarts ("I was kind of, um, having, um..."), and self-corrections ("me, I'm not, I'm not someone that really travels a lot"). These elements indicate that speakers are thinking in real time, shaping their responses dynamically rather than delivering pre-formulated statements.

Additionally, the presence of non-verbal elements like coughing further highlights the informal and unpolished nature of natural speech. These interruptions, along with ungrammatical structures ("spend a little bit of my, um, little age"), demonstrate how everyday conversations are shaped by real-world spontaneity rather than strict adherence to linguistic rules.

Another important aspect is interactivity—the conversation is collaborative, with speakers responding to each other in a way that builds on shared experiences ("And just like you, I don’t really travel a lot"). This turn-taking and mirroring of ideas contribute to the natural flow of conversation, reinforcing the social and contextual nature of spoken interactions.

\subsubsection{Long-Context Dependency}
The transcript also demonstrates long-context dependency in spontaneous conversations, often referred to as the "callback" effect in dialogue dynamics. This occurs when a speaker references or builds upon something mentioned earlier in the conversation, even after a long exchange.

For example, Spk2 initially states, "I'm not someone that really travels a lot." Much later, Spk1 recalls and aligns with this sentiment, saying, "And just like you, I don’t really travel a lot." This callback reinforces conversational cohesion, showing that speakers retain and integrate past dialogue into later turns. This phenomenon is crucial in context-aware speech processing, as it highlights how natural conversations maintain semantic continuity over extended interactions, even with intervening topics.

Such long-range dependencies pose challenges for automatic speech recognition (ASR) and dialogue systems, as models need to track and retrieve relevant past information dynamically rather than processing speech as isolated utterances.


\subsection{Benchmark}
\subsubsection{Automatic Speech Recognition}
To assess the quality of our dataset and establish baseline performance, we randomly sampled 1.95 hours of speech from various conversations. These segments were manually aligned and transcribed using Label Studio, a collaborative annotation platform, ensuring high-quality reference transcripts.

We evaluated two state-of-the-art speech recognition models: Whisper-large-v3 and SeamlessM4T-large. Whisper-large-v3 is a multi-task encoder-decoder model with 1.6B parameters trained on a diverse collection of speech data \cite{radford2022robustspeechrecognitionlargescale}, demonstrating strong generalization across different domains. SeamlessM4T-large is a massively multilingual and multimodal model with 2.3B parameters designed for speech and text translation and recognition \cite{seamlessm4t}, making it particularly relevant for evaluating spontaneous, conversational speech.

\begin{table}[h]
\caption{Baseline ASR performance on a 1.95-hour subset of our dataset.}
\centering
\begin{tabular}{l c}
\toprule
\textbf{Model} & \textbf{WER} \\
\midrule
Whisper-large-v3 & 0.31 \\
\midrule
SeamlessM4T-large & 0.53 \\
\bottomrule
\end{tabular}
\label{tab:benchmark}
\end{table}

We computed word error rate (WER) using Whisper’s built-in English normalizer for both references and hypotheses. The results, shown in Table~\ref{tab:benchmark}, indicate that Whisper-large-v3 achieves a WER of 0.31, while SeamlessM4T-large performs worse with a WER of 0.53. These relatively high error rates highlight the challenges of spontaneous, conversational speech, which includes disfluencies, informal phrasing, and long-context dependencies—characteristics that are underrepresented in the training data of these models. This suggests that our dataset is a valuable resource for improving ASR models, particularly for real-world conversational speech.

\subsubsection{Diarization}
We evaluated the 102 hours of publicly available conversations by 157 distinct speakers on the diarization task. Each conversation consists of two time-aligned audio files, one from each participant’s end. We combined the two files into a single-channel recording, which was then used for diarization evaluation.

WhisperX \cite{bain2022whisperx} is an ASR model built on Whisper that employs forced alignment with wav2vec2 \cite{wav2vec2} to generate word-level timestamps. To create the ground truth, speech segments were identified using WhisperX-generated timestamps on each participant’s recording independently, for every conversation. These timestamps were mapped onto the combined conversation to be used as reference labels. 

For the benchmark of diarization performance, we used two state-of-the-art diarization models: pyannote.audio-v3.1\cite{pyannote1, pyannote2} and the NeMo Cascaded Speaker Diarization System\cite{nemo}. Pyannote-audio v3.1 is a fully automatic diarization system that separates speakers without relying on an external VAD model, and is widely adopted in diarization tasks and used for benchmarking. The model uses a combination of EEND \cite{Fujita2019Interspeech} diarization and unsupervised clustering.

The NeMo framework, developed by NVIDIA, includes a neural cascaded diarization architecture combining specialized components for each sub-task. Our benchmark uses TitaNet-Large\cite{koluguri2021titanetneuralmodelspeaker}, a speaker embedding model trained on thousands of hours of speech, and MarbleNet\cite{jia2021marblenetdeep1dtimechannel}, a VAD model trained on over thousands of hours of multi-language speech data. Together, these components enable robust speaker segmentation and clustering, making the NeMo Neural Diarizer well suited for evaluating this dataset.

\begin{figure}[htbp]
  \centering
    \includegraphics[width=0.5\textwidth, clip]{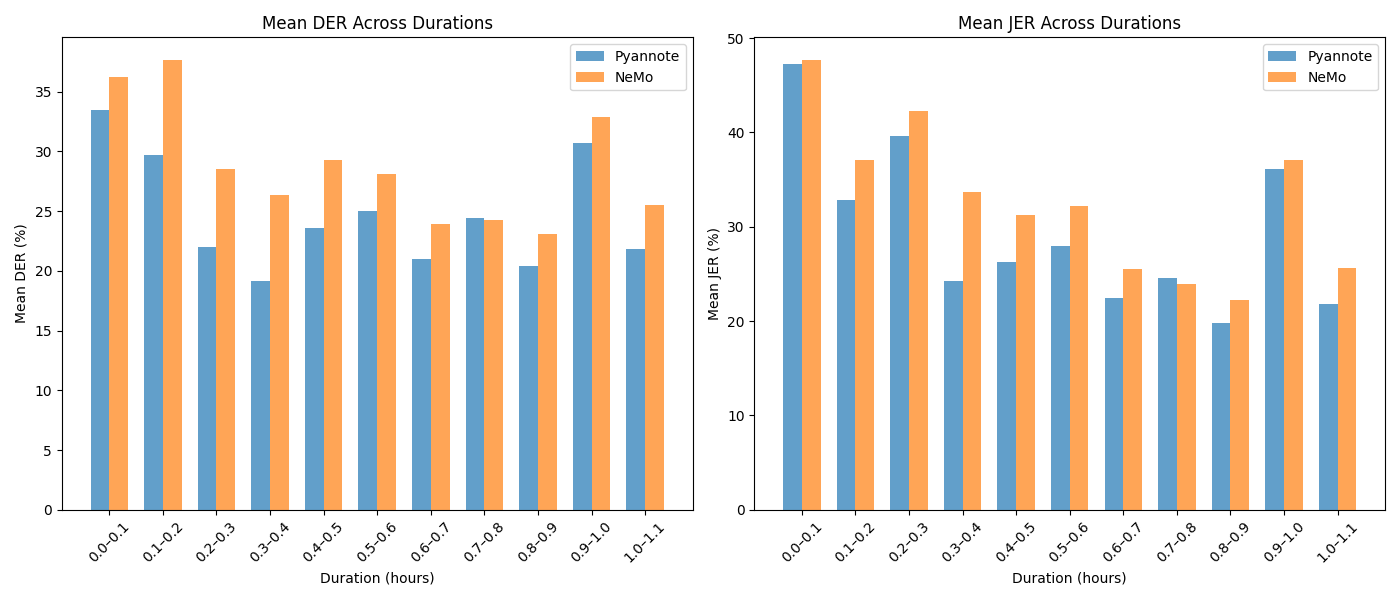}
  \caption{Histogram of Diarization Evaluation Results}

  \label{fig:diarization}
\end{figure}

We measure Diarization Error Rate (DER) and Jaccard Error Rate (JER) computed without a forgiveness collar and with regions of overlapped-speech included in the evaluation. The results metrics can be seen in Table \ref{tab:diarization}  as well as the histogram of error rates across recordings in Fig. \ref{fig:diarization}. The histogram shows the mean DER and JER across bins of recording durations, enabling the evaluation of diarization performance that is related to length of the recording. The means are over each bin only, the number of files in each bin affect the overall error rates seen in Table \ref{tab:diarization}.

\begin{table}[ht]
\centering
\caption{Diarization performance metrics (DER and JER) for Pyannote and NeMo models. Reported are the mean, median, 5th percentile, and 95th percentile values across all recordings.}
\label{tab:diarization}
\begin{tabular}{lcccc}
\toprule
\textbf{Model} & \textbf{Metric} & \textbf{Mean (\%)} & \textbf{Median (\%)} & \textbf{5\% / 95\% (\%)} \\
\midrule
\multirow{2}{*}{Pyannote} & DER & 25.93 & 23.24 & 16.80 / 51.61 \\
                          & JER & 29.63 & 24.59 & 15.48 / 57.20 \\
\midrule
\multirow{2}{*}{NeMo}     & DER & 30.77 & 28.23 & 16.80 / 53.48 \\
                          & JER & 33.87 & 30.85 & 17.62 / 60.38 \\
\bottomrule
\end{tabular}
\end{table}

The relatively high error rates on our dataset suggest an increased difficulty of diarization on conversational speech with disfluencies, overlaps, and frequent turn-taking dynamics that are underrepresented in existing resources. The natural setting of recordings, where each participant records from their own device in their own environment, also introduces new obstacles for the system. These results indicate that our dataset presents real-world challenges in diarization systems and can be a suitable and valuable resource in the speech community to improve the accuracy and robustness of diarization models for conversational speech. 

\section{Conclusion}
In this paper, we introduced a new dataset of spontaneous conversational speech, totaling 200 hours of recorded audio, with 158 hours identified as actual speech. The dataset captures the natural flow of dialogue, including disfluencies, pauses, and speaker interactions, primarily between acquaintances. This makes it a valuable resource for developing more robust downstream task systems (ASR, diarization, topic detection, among others) capable of handling real-world conversational scenarios. This first release comprises a subset of 102 hours of audio, publicly available. 

Evaluation of the baseline models on a randomly sampled subset of our dataset highlights the difficulties ASR and diarization systems face with spontaneous speech. This result underscores the need for datasets like ours to train models that can better handle natural conversations, disfluencies, and dynamic speech patterns. Our work provides a foundation for future research aimed at improving downstream task systems, and we plan to expand the dataset and evaluate additional models to enhance performance further.

\clearpage
\bibliographystyle{IEEEtran}
\bibliography{mybib}

\end{document}